\documentclass[10pt,twocolumn,letterpaper]{article}

\usepackage{3dv}
\usepackage{times}
\usepackage{epsfig}
\usepackage{graphicx}
\usepackage{amsmath}
\usepackage{amssymb}

\usepackage{algorithm}
\usepackage{algpseudocode}

\usepackage[colorlinks,
            linkcolor=red,
            anchorcolor=blue,
            citecolor=green
            ]{hyperref}
\usepackage{multirow}
\usepackage{booktabs}
\usepackage{threeparttable}
\usepackage{biblatex}
\threedvfinalcopy 

\def\threedvPaperID{****} 

\ifthreedvfinal\fi
\begin{document}

\title{Depth-Independent Depth Completion via Least Square Estimation}

\author{Xianze Fang, Yunkai Wang, Zexi Chen, Yue Wang, Rong Xiong\\
Zhejiang University\\
{\tt\small \{fangxzzju,wangyunkai,chenzexi,ywang24,rxiong\}@zju.edu.cn}
}

\maketitle
\thispagestyle{empty}

\begin{abstract}
The depth completion task\cite{hu2022deep} aims to complete a per-pixel dense depth map from a sparse depth map. In this paper, we propose an efficient least square based depth-independent method to complete the sparse depth map utilizing the RGB image and the sparse depth map in two independent stages. In this way can we decouple the neural network and the sparse depth input, so that when some features of the sparse depth map change, such as the sparsity, our method can still produce a promising result. Moreover, due to the positional encoding and linear procession in our pipeline, we can easily produce a super-resolution dense depth map of high quality. We also test the generalization of our method on different datasets compared to some state-of-the-art algorithms. Experiments on the benchmark show that our method produces competitive performance.

\end{abstract}

\section{Introduction}
Depth estimation \cite{ming2021deep} is an important task in many areas, such as robotics and virtual reality. Although existing depth sensors contribute a lot to 3D perception tasks, they still have some limitations \cite{huang2019hms, zhang2018deep} in directly obtaining a reliable dense depth map in both indoor and outdoor environments. For example, LiDARs can produce accurate distance measurements but their measurements are sparse, while structured light sensors can produce dense depth maps but are constrained to indoor environments. The depth completion task aims at predicting a reliable dense depth map utilizing sparse depth measurements, which can be easily obtained by the existing sensors.

\begin{figure}[tb]
  \centering
  \includegraphics[width=\linewidth]{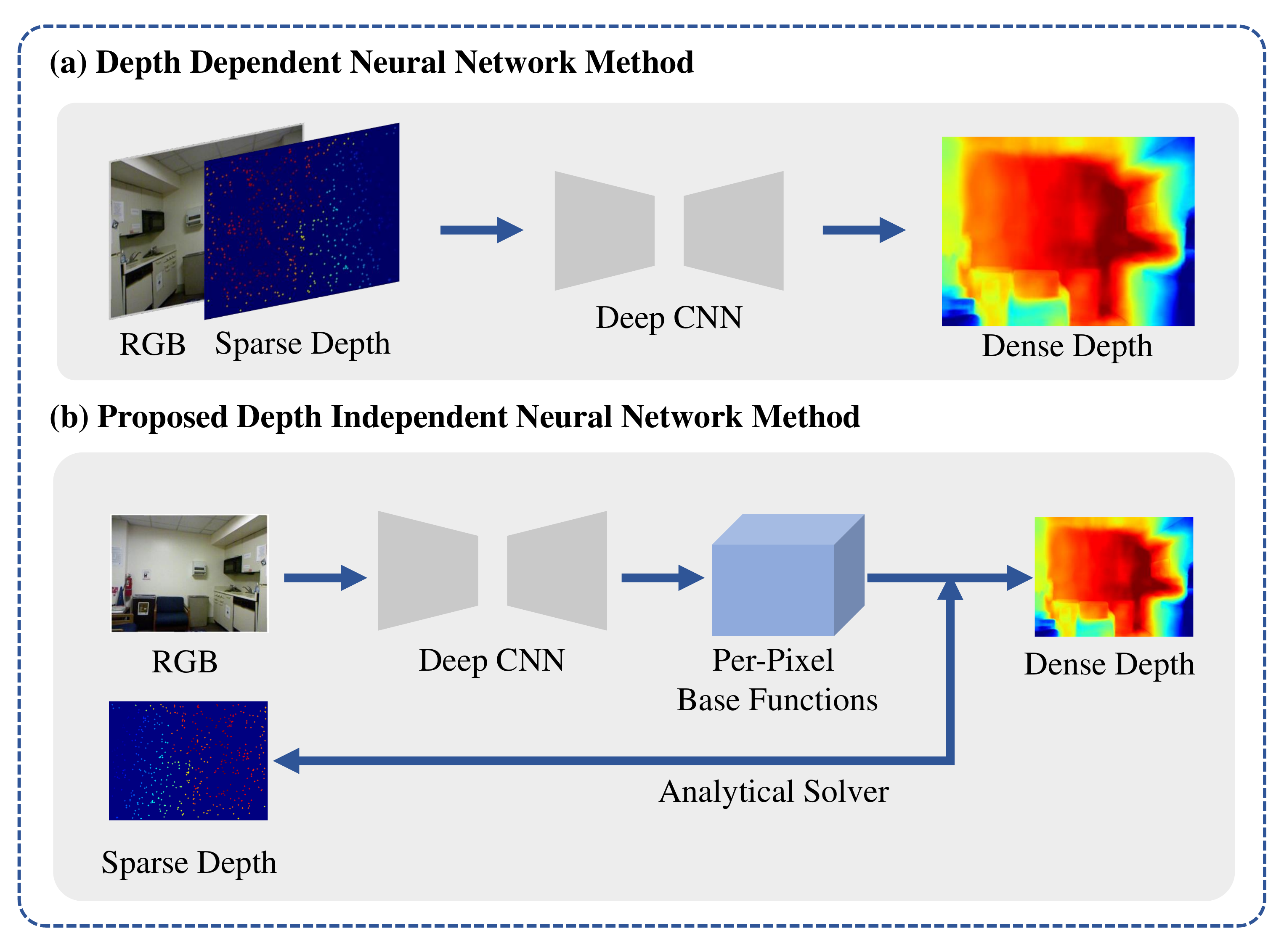}
  \caption{The pipeline of depth-dependent and our proposed depth-independent methods for depth completion. Our proposed method does not use the sparse depth map as the input of the neural network and uses a least square algorithm to predict the dense depth map.}
  \label{introduce_pic}
  \vspace{-10pt}
\end{figure}
The main challenge of the depth completion task is the extreme sparsity of the depth measurements (e.g. less than $1\%$ pixels have depth measurements). Recently neural network has achieved great success in many image processing areas including the depth completion task. In all these learning-based depth completion algorithms, we can divide them into two main categories which are shown in Fig.~\ref{introduce_pic}. The first class is a depth-dependent neural network method \cite{ma2018sparse, Qu_2020_WACV, jaritz2018sparse, 8793637}. The pixels without valid depth are set to a constant number such as 0. The RGB image is concatenated with the sparse depth map as the input of the deep neural network. These methods can be used in an end-to-end manner, but their neural networks need to resolve the ambiguity between valid and invalid depth measurements, and to learn the features of the sparse depth map such as the sparsity, the noise of the sparse measurements and the distributions of the sparse points. These factors make themselves less flexible for the changing of the sparse depth map. Moreover, the learning space of their neural networks should be large to cover the sparse depth map inputs, which not only increases the parameters of their networks but also limits the generalization abilities.

Another kind of algorithm uses depth-independent neural networks which do not directly use the sparse depth map as the input of the neural networks \cite{8869936, cheng2020cspn++}. Liu. \textit{et al.} \cite{liu2021learning} are one of the pioneers who propose a depth-independent \textit{Learning Steering Kernel (LSK)} method to complete the sparse depth. However, the regression kernels they use are inflexible and too simple to model the complicated images. Therefore, they need to use a depth-dependent CNN to finetune the predicted depth map in addition. Followed \textit{LSK}, our proposed method is shown in Fig.~\ref{introduce_pic} (b). We use the neural network only to extract the features of the guided RGB image, and then use a analytical solver to fit the sparse depth measurements. In this way when the sparsity or other features of the sparse depth map change, the output of the neural network will not be impacted at all. In addition, to improve the performance and be able to upsample the dense depth map, we add positional encoding layers to the RGB features. Compared with kernels in \textit{LSK}, our base functions are more flexible. The parameter number of our neural network is small because the network is only used for RGB image feature extraction. We evaluate our method on public datasets compared with other methods. Experiments on benchmarks show that our method outperforms some previous works in accuracy and runtime. In conclusion, our contributions are summarized as follows:

\begin{itemize}
\item[1)] The neural network we use is depth-independent which makes our model more flexible for the feature changing of the sparse depth map input. We propose a base function representation for images which has greater expressiveness than regression kernels.
\item[2)] With the extra positional information of pixels added, we can get super-resolution dense depth maps of high quality.
\item[3)] Our method is light-weighted and fast compared with other methods. With about $1\%$ number of SOTA's network parameters,we achieve competitive results on generalization experiments.
    
    
\end{itemize}

We organize our paper as follows: Sec.~\ref{sec:related_works} will review the related works of the related areas. We will explain our method in detail in Sec.~\ref{sec:method}. Sec.~\ref{sec:experiment} will present our experiment results in detail. Finally, we will conclude our paper in Sec.~\ref{sec:conclusion}.

\begin{figure*}[htb]
  \centering
  \includegraphics[width=0.9\textwidth]{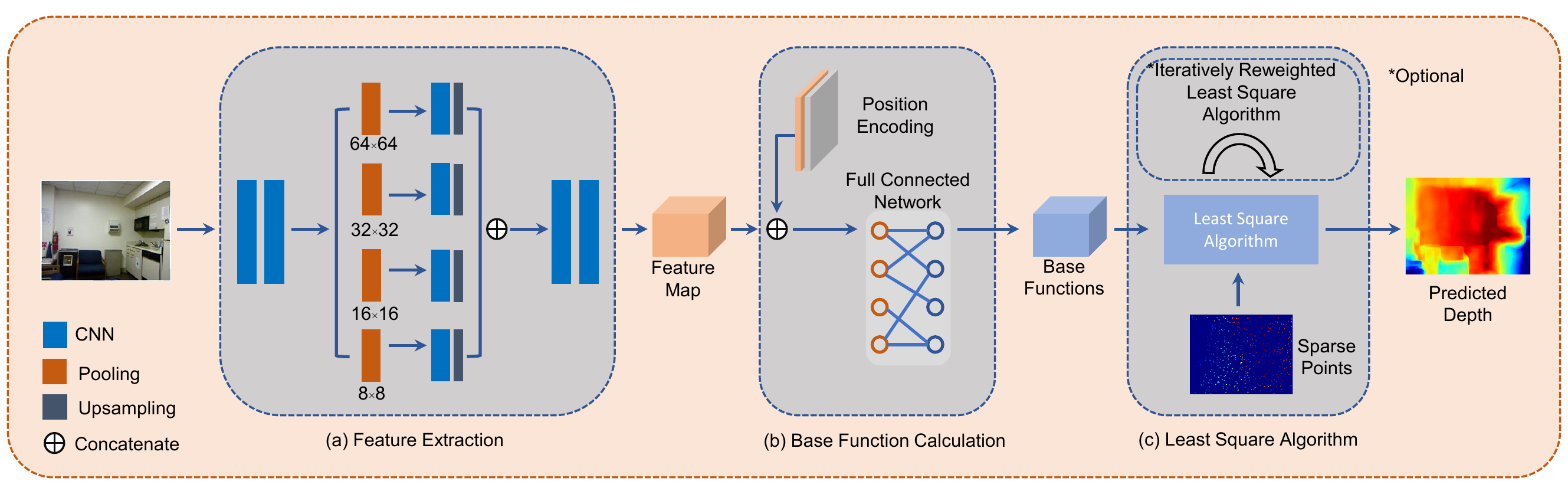}
  \caption{The whole pipeline of our method. We divide our algorithm into three parts. (a) RGB image feature extraction via CNN. (b) Positional encoding and base function estimation. (c) The least square optimization fitting the base functions and the sparse depth measurements.}
  \label{structure}
\end{figure*}

\section{Related Work}
\label{sec:related_works}

\subsection{Monocular Depth Estimation}
Many previous works have been done on predicting the depth map from a monocular RGB image. Early works \cite{saxena2005learning} \cite{saxena2008make3d} rely on hand-crafted features and probabilistic graphical models. Recently many state-of-the-art methods based on supervised models are proposed and make good use of convolutional neural networks. These models are always in end-to-end manner. In addition, some other works \cite{vandenhende2020mti, xu2018pad} use multi-task models to synchronously predict the dense depth maps and segmentation results, which can benefit the performance of each other. Some methods also predict the uncertainty map at the same time, which is used to finetune the predicted depth map. Bhutani \textit{et al}. \cite{bhutani2020unsupervised} proposes unsupervised methods which also demonstrate promising results. 

Although many monocular depth estimation methods reach promising results, the task that predicting the dense depth map only from a monocular RGB image is highly ill-posed. Their generalization abilities are limited.

\subsection{Depth Completion}
With an extra sparse depth map known, the depth completion task can predict more reliable dense depth maps. The sparse depth measurements can be obtained from visual-SLAM algorithms like ORB-SLAM \cite{mur2015orb} and Vins-MONO \cite{qin2018vins}, or directly obtained from the distance sensors like LiDARs and structured light sensors. Many methods use end-to-end neural networks to complete the depth map from sparse to dense. Ma \textit{et al.} \cite{ma2018sparse} get a promising result using an end-to-end convolutional neural network to proceed with the concatenation of RGB image and the sparse depth map. As a depth-dependent method it runs fast but when the features of the input sparse depth change, the performance is not good. Tang \textit{et al.} \cite{tang2018ba} prove that depth map can be represented using a series of basis. Takenda \textit{et al.} \cite{takeda2007kernel} propose a non-parametric statistics estimation method to interpolate the sparse depth map, but the steering kernels it use is hand-crafted and inflexible. Liu \textit{et al.} \cite{liu2021learning} promote the algorithm using learned steering kernels to replace the hand-crafted steering kernels, which significantly improves the performance. Spatial propagation is a traditional interpolated algorithm. Liu S. \textit{et al.} \cite{liu2017learning} used a neural network to predict the affinity matrix of the image and use the affinity matrix to iteratively finetune the output which can be used in many vision tasks. Cheng \textit{et al.} \cite{cheng2018depth} applied the spatial propagation algorithm in the depth completion task. Park \textit{et al.} \cite{park2020non} promoted the convolutional spatial propagation network by using a flexible set of neighbor points to replace the fixed set. By doing this, it can learn the affinities of non-local points of the image. The spatial propagation network methods need an iterative regression process and their networks are large, making them inefficient.

In this paper, we propose a differentiable depth-independent network which decouples the RGB image and the sparse depth map. Our model is flexible for the changing of the sparse depth map, and we can upsample a super-resolution dense depth map of high quality. 

\section{Method}
\label{sec:method}
\subsection{Motivation}
Our goal is to predict a dense depth map utilizing the RGB images and the sparse depth maps in a decoupling way. \cite{Qu_2020_WACV, tang2018ba} have proven that depth map can be represented using a series of basis, which we call \textit{base functions}. Following this thought, we use neural network to estimate the base functions of the RGB image. Then, like the final layer of a normal neural network, we regard the dense depth map as the linear combination of the \textit{base functions}. Therefore we propose a least square estimation method to fit the base functions and the sparse depth map. We will describe each part in detail.

\subsection{Neural Network Architecture} 


\noindent\textbf{Feature Extraction.} 
Traditional CNN detector shows excellent performance in detecting the image regions which have rich object context information. But in some ill-posed regions which have few textures, it's difficult to detect the features. We follow the idea in \cite{chang2018pyramid} by extracting the features from the RGB image with a pyramid-like CNN. We use $4$ different size average pooling blocks in the network to extract different level features. Then we concatenate them and output the feature map of the RGB image. We can describe this part as follows:
\begin{equation}
    h=\phi(\mathbf{c})
\end{equation}
where $\phi$ refers to the feature extraction network, and $\mathbf{c}$ refers to the RGB value of each pixel.

\noindent\textbf{Positional Encoding.} 
Traditionally, we focus on the color information of the pixel but ignore the coordinate information. According to \cite{liu2018intriguing}, adding coordinate layers to the feature map can improve the performance in many applications including depth prediction. Moreover, according to \cite{czarnowski2020deepfactors}, mapping the coordinate to high dimensions contributes to representing the high-frequency features of the image. These techniques enable better fitting of the depth image. The encoding function we used is as follows:

\begin{equation}
\label{equ:pos_endode}
    \begin{split}
        \gamma{(\textbf{p})}=[x, x, y, y, sin(2^0\pi x), cos(2^0\pi x), \\sin(2^0\pi y), cos(2^0\pi y), ...,sin(2^{E-2}\pi x),\\ cos(2^{E-2}\pi x), sin(2^{E-2}\pi y), cos(2^{E-2}\pi y)]
    \end{split}
\end{equation}
where $ E $ is the number of dimensions of positional encoding, $ \mathbf{p}=(x, y) $ denotes pixel position, $ x $ denotes coordinates along the width, and $ y $ along the height. The output of the positional encoding $\gamma{(\textbf{p})}$ is then concatenated with the feature map as the input of the base functions estimation part.

\noindent\textbf{Base Functions Estimation.}
\label{sec:base_function_estimate}
We concatenate the feature map and the encoded positional map as the input of a \textit{fully-connected neural network (FCNN)}. The outputs of the \textit{FCNN} are vectors that are one-to-one corresponding to the pixels. We call these vectors the \textit{base functions}. For each pixel, we have:
\begin{equation}
 \label{eqa:base_function}
    \mathbf{f}_k=\psi(\phi(\mathbf{c}), \gamma(\mathbf{p}))
\end{equation}
where $k$ refers to the index of each pixel, $\psi$ refers to the \textit{FCNN}, and $\mathbf{f}_k$ is a $N\times1$ vector.


\subsection{Least Square Optimization and IRLS Algorithm}
\noindent\textbf{Solving the Least Square Problem.}
After feature extraction and base function estimation, we get a base vector $\mathbf{f_k}\in \mathbb{C}^{N\times1}$ of each pixel, which is called base function. Assuming there are $m$ pixels in the RGB picture, of which $n$ pixels have a known depth value. The known sparse depth is $\mathbf{D}\in\mathbb{C}^{n\times1}$. For each pixel, we have the equation:

\begin{equation}
    d_k=w_0 + \sum_{i=1}^{N}{w_i \cdot \mathbf{f}_k[i]}\\
\end{equation}
or we can write the above equation in matrix forms as:

\begin{gather}
\label{eqa:lse}
     d_k=\mathbf{f}_k\cdot\mathbf{W^T}\notag\\
        \mathbf{f}_k=[f_0,f_1,...,f_N],f_0=1\notag\\
        \mathbf{W}=[w_0,w_1,...,w_N]
\end{gather}

For the whole picture, we use $ \mathbf{F}_{a}\in\mathbb{C}^{m*(N+1)},m>(N+1) $ to represent the base functions of all the pixels and $ \mathbf{F}_{s}\in\mathbb{C}^{n*(N+1)},n>(N+1) $ for depth-known pixels, while $ \mathbf{D}_a\in \mathbb{C}^{m*1} $ for depth values for all the pixels and $ \mathbf{D}_s\in \mathbb{C}^{n*1} $ for known depth values. We have the equations below:
\begin{gather}
        \mathbf{D}_a=\mathbf{F}_a\cdot\mathbf{W}^T\\
        \label{functions_to_be_solve}
        \mathbf{D}_s=\mathbf{F}_s\cdot\mathbf{W}^T
\end{gather}
As described in Sec.~\ref{sec:base_function_estimate}, we already have the $\mathbf{F}_s$ which is sampled from $\mathbf{F}_a$ according to the known depth pixels. So now our task is to solve Eqn.~\ref{functions_to_be_solve} which is an over-defined matrix equation. This equation can also be written as a least square estimation problem as follows:
\begin{equation}
    \mathop{\arg\min}_{\mathbf{\overline{W}}} \| \mathbf{D}_{s}-(\mathbf{F}_s\cdot\mathbf{\overline{W}}^T)\|.
\end{equation}

\begin{algorithm}[h]
  \caption{SVD-decomposition Solving Least Square Problem}
  \label{least square method}
  \begin{algorithmic}[1]
    \Require
    $
      \mathbf{D}_s\in\mathbb{C}^{n\times1}, \mathbf{F}_s\in\mathbb{C}^{n\times(N+1)}
    $
    \Ensure
      $ \mathop{\arg\min}_{\mathbf{\overline{W}}} \| \mathbf{D}_{s}-(\mathbf{F}_s\cdot\mathbf{\overline{W}}^T)\|. $;
    \If{$ rank(\mathbf{F}_s)=r=(N+1)<n $}
    \State $ SVD(\mathbf{F}_s)=(\mathbf{U}_1,\mathbf{U}_2)\begin{pmatrix}\Sigma_1\\0 \end{pmatrix}\mathbf{V}^H $
    \State $ \mathbf{\overline{W}}=\mathbf{V}\Sigma^{-1}_1\mathbf{U}^H_1\mathbf{D}_s $
    \Else({$ rank(\mathbf{F}_s)=r<(N+1)<n $})
    \State $ SVD(\mathbf{F}_s)=(\mathbf{U}_1,\mathbf{U}_2)\begin{pmatrix}\Sigma_1&0\\0&0\end{pmatrix}\begin{pmatrix}\mathbf{V}_1^H\\ \mathbf{V}_2^H \end{pmatrix} $
    \State $ \mathbf{\overline{W}}=\mathbf{V}_1\Sigma^{-1}_1\mathbf{U}^H_1\mathbf{D}_s $
    \EndIf
  \end{algorithmic}
\end{algorithm}

\begin{algorithm}[h]
  \caption{Iterative Re-weighted Least Square Algorithm}
  \label{IRLS method}
  \begin{algorithmic}[1]
    \Require
    $
      \mathbf{D}_s\in\mathbb{C}^{n\times1}, \mathbf{F}_s\in\mathbb{C}^{n\times(N+1)}
    $
    \Ensure
      $ \mathbf{\overline{W}}_{init}[w_i], (i=0,1,2,...,N) $;
    \State $ \mathbf{W}=\mathbf{\overline{W}}_{init}=(\mathbf{F_s} \cdot \mathbf{F_s^T})^{-1}\cdot (\mathbf{F_s^T}\cdot \mathbf{D_s}) $
    \While{not meet the stopping criterion}
 
        \State  $ \mathbf{e}=abs(\mathbf{D_s}-\mathbf{F_s}\cdot \mathbf{W}) $
        \State  $ \mathbf{e'}=({max}(threshold,\mathbf{e}))^{-1} $
        \State  $ \mathbf{Weight}=diag(\mathbf{e'})$
        \State $ \mathbf{W}=(\mathbf{Weight}\cdot \mathbf{F_s} \cdot \mathbf{F_s^T})^{-1}\cdot (\mathbf{Weight}\cdot\mathbf{F_s^T}\cdot \mathbf{D_s}) $
    \EndWhile
  \end{algorithmic}
\end{algorithm}

We can use \textit{Moore-Penrose} inverse of matrix $ \mathbf{F}_s $ to solve the least square regression problem. According to \textit{Moore-Penrose} inverse theory, we have:
\begin{equation}
    \mathbf{W}^T=\mathbf{F}_s^+\cdot\mathbf{D}_s=(\mathbf{F}_s\cdot\mathbf{F}_s^T)^{-1}\cdot\mathbf{F}_s^T\cdot\mathbf{D}_s
\end{equation}
In the training process, we should keep the least square estimation differentiable so we can train the model end-to-end. But in the practical try, we find that the inverse function in Pytorch may easily reach a numerical instability condition because the dimension of the matrix $ \mathbf{F}_s\cdot\mathbf{F}_s^T $ is too big for the algorithm to converge so the matrix may be treated as a singular matrix. So we use the SVD-decomposition algorithm to solve the least square estimation problem which is more numerically stable in our training process. The process is shown in Alg.~\ref{least square method}.

\noindent\textbf{IRLS Algorithm.}
In real-world applications, there might be some noise in the sparse depth measurements. In our pipeline, we can easily apply an iterative \textit{re-weighted least square algorithm (IRLS)} to reduce the influence of the noise. The algorithm is shown in Alg.~\ref{IRLS method}. The \textit{IRLS} algorithm is optional in the whole pipeline.

\subsection{Loss Functions}
In this part, we will explain the loss functions we used in training. Our training loss consists of three parts. We will explain each part in detail next.

\begin{itemize}
\item [1)] \textbf{Loss on predicted depth.} 
We directly estimate the error between the predicted depth and ground truth depth:
\begin{equation}
    \ell_{direct}=\| \mathbf{D_{gt}}-\mathbf{D_{predict}} \|_t
\end{equation}
When $ t=1 $ it represents $ \ell_1 $ loss and $ t=2 $ represents $ \ell_2 $ loss. We use both $ \ell_1 $ loss in our method in our training.

\item [2)] \textbf{Loss on depth gradient.} 
Previous work has shown that applying depth gradient loss in training can significantly improve the performance. The gradient of the depth map contains information about the edges of objects. Therefore this loss supervises the boundaries in predictions and makes them sharper.
\begin{equation}
    \ell_{grad}=\| \nabla \mathbf{D_{gt}} - \nabla \mathbf{D_{predict}} \|_1
\end{equation}

\item [3)] \textbf{Loss on regularization}
We regard the depth value of each pixel as a weighted sum of base functions. To make the most use of each base function and avoid over-fitting, we apply a regularization loss function in training.
\begin{equation}
    \ell_{regular}=\| \sum_{i=0}^{N}w_{i} \|_1
\end{equation}

\end{itemize}

The total loss is the weighted combination of the above loss functions:
\begin{equation}
    \mathcal{L}=\alpha \ell_{direct}+\beta \ell_{grad} + \gamma \ell_{regular}
\end{equation}
where $\alpha,\beta,\gamma$ are the hyper-parameters set manually.

\begin{table}
\centering
\begin{threeparttable}[b]
\caption{Depth Completion Benchmark on NYUDepthv2}
\label{nyudv2_result}
\begin{tabular}{c@{\ \ \ \ }c@{\ \ \ \ }c@{\ \ \ \ }c@{\ \ \ \ }c@{\ \ \ \ }c}
\toprule
\multirow{2}{*}{Method\tnote{1}} &  \multicolumn{2}{c}{Error($\downarrow$)}  & \multicolumn{3}{c}{Accuracy($\uparrow$)}\\
\cmidrule(lr){2-3}
\cmidrule(lr){4-6}
 & RMSE&  REL&  $ \delta_{1.25} $ & $ \delta_{1.25^2} $ & $ \delta_{1.25^3} $\\
 \midrule
 S2D&  0.230&  0.044&  97.12&  99.37&  99.83\\
 SK&  0.221&  0.043&  96.87&  99.51& 99.83\\
 LSK\tnote{2}&  0.198&  0.034&  97.73&  99.80& 99.89 \\
 Ours& \textbf{0.148}& \textbf{0.031}& \textbf{98.72} & \textbf{99.84} & \textbf{99.95} \\
 \bottomrule
\end{tabular}
   \begin{tablenotes}
     \item[1] \footnotesize{All take 500 sampled points.}
     \item[2] \footnotesize{The result of Learning Steering Kernel part.}
   \end{tablenotes}
\end{threeparttable}
\end{table}

\section{Experiments}
\label{sec:experiment}

In this section, we will first introduce our results on the NYUDepthv2 dataset compared with other methods. Secondly, we will show the ablation studies of our method. Thirdly we will show the ability of our method to deal with the feature changes of the input sparse depth map. Then we will show the ability of our method to super-resolution the dense depth map. Last but not least, we will compare the generalization ability of our method with other algorithms.

\subsection{Experiments on Benchmark}
\noindent\textbf{NYUDepthv2 Dataset.}
The NYUDepthv2 dataset \cite{Silberman:ECCV12} contains about $407k$ RGB frames with depth maps in $464$ different indoor scenes. There are $1449$ labeled frames in it whose depth is labeled densely. A small set including $654$ images is officially chosen for the benchmark. The resolution of the raw image in NYUDepthv2 is $ 480\times640 $. We first downsample the image to half the resolution and then center-crop them to $ 228\times304 $ in training and evaluation.

\noindent\textbf{Experiment Setup.}
We implement our method using PyTorch 1.8 with CUDA 10.2. We train our model on the NYUDepthv2 dataset using an Nvidia 2080Ti GPU. The weights of the network are initialized randomly. We use a small batch size $2$ and train the model for $100$ epochs. We set the learning rate to $1e-4$ for the first $30$ epochs and then reduce it to $1e-5$. We set the weight decay to $5e-4$ for regularization.

\begin{figure}
  \centering
  \includegraphics[width=0.46\textwidth]{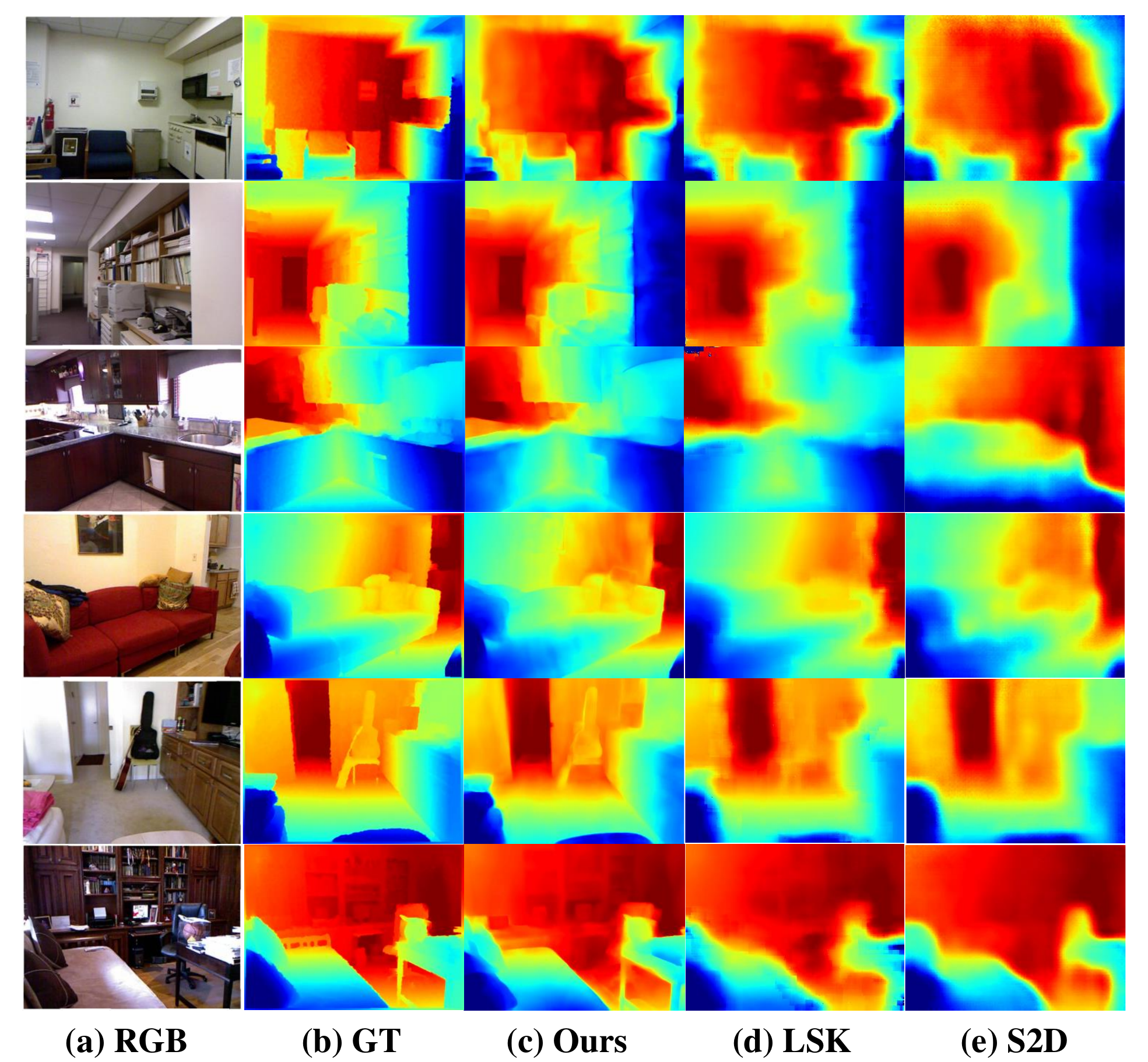}
  \caption{Compare results of experiments on NYUDepthv2 benchmark. In all methods the sampled number in sparse depth map is $1000$.}
  \label{nyudv2_pic}
  \vspace{-10pt}
\end{figure}

\noindent\textbf{Evaluation Metrics.}
We adopt the standard evaluation metrics. 
Assuming that the ground truth depth value at position $ \mathbf{x_{i}} $ is $d_i=D_i(\mathbf{x_i})$, the predicted depth value at position$\mathbf{x_i}$ is $\hat{d_i}=\hat{D_i}(\mathbf{x_i})$. The evaluation metrics are specified as follows:

1) \textit{Root Mean Squared Error (RMSE)}:
\begin{equation}
    RMSE=\sqrt{\frac{1}{M}\sum_{i}{(\hat{d_i}-d_i)^2}} 
\end{equation}\par
2) \textit{Mean Absolute Relative Error (REL)}:
\begin{equation}
    REL=\frac{1}{M}\sum_{i}{\frac{\| \hat{d_i}-d_i \|}{\| d_i \|}}
\end{equation}\par
3) Threshold ($\delta \in[1.25, 1.25^2, 1.25^3] $): 
\begin{equation}
Percentage~of~\hat{d_i} ~s.t.~ max(\frac{\hat{d_i}}{d_i}, \frac{d_i}{\hat{d_i}})<\delta
\end{equation}\par

\noindent\textbf{Benchmark Evaluation.} 
We evaluate our model on the NYUDepthv2 benchmark. Tab.~\ref{nyudv2_result} shows the comparison results on the NYUDepthv2 benchmark of accuracy. Our model outperforms \textit{Sparse-to-Dense (S2D)}, the interpolated result of \textit{Steering Kernels (SK)} and the interpolated result of \textit{Learning Steering Kernels (LSK)}. 

\subsection{Ablation Study}

\begin{figure*}[htb]
  \centering
  \includegraphics[width=\textwidth]{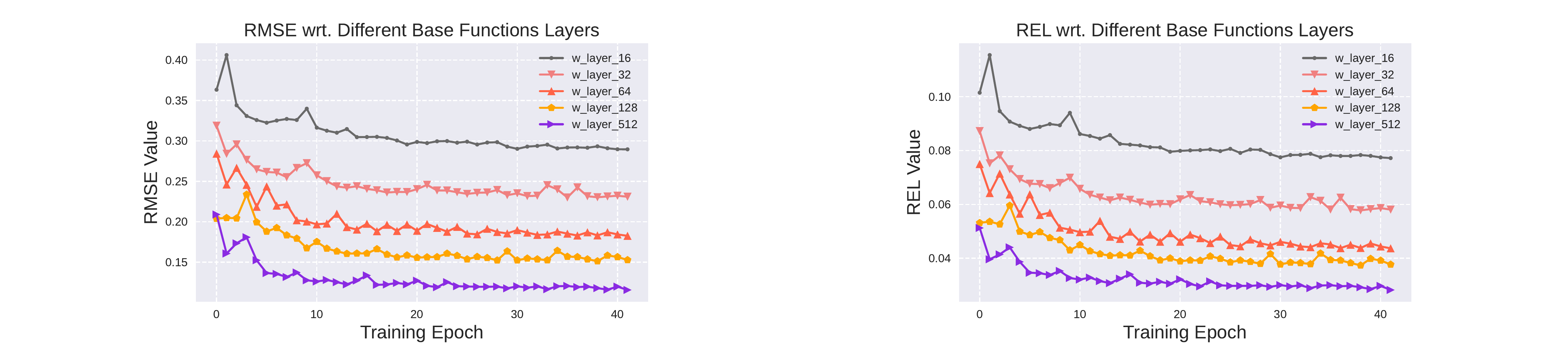}
  \caption{Ablation study on different dimensions of base functions. Except for the layer, we use the same settings to train our model from the beginning. And we evaluate each model after each epoch. The REL and RMSE is evaluate in the testing subset of NYUDepthv2 dataset.}
  \label{wlayer}
 \vspace{-10pt}
\end{figure*}

\begin{figure*}[htb]
  \centering
  \includegraphics[width=\textwidth]{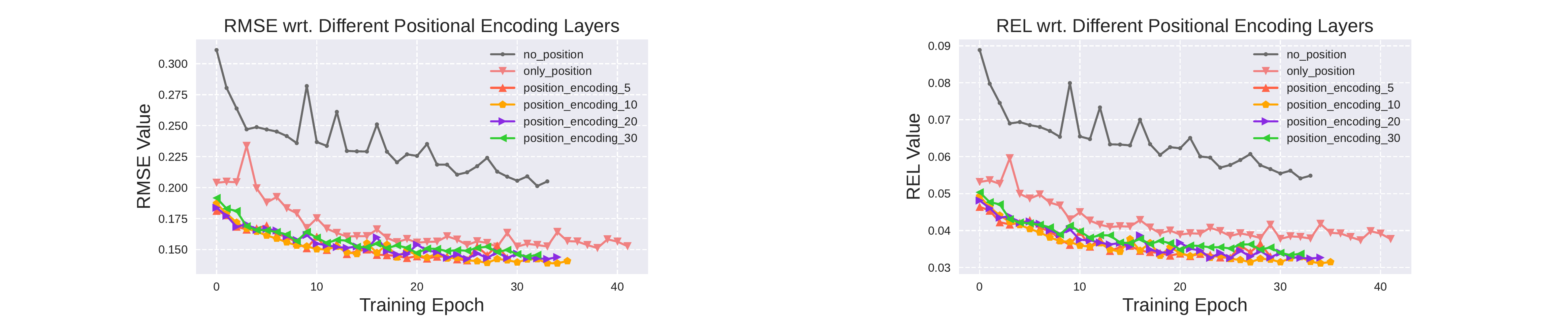}
  \caption{Ablation study on different layers of positional encoding. Except for the positional encoding part, we use the same settings to train our model from the begining. And we evaluate each model after each epoch. The REL and RMSE is evaluate in the testing subset of NYUDepthv2 dataset.}
  \label{pe}
  \vspace{-10pt}
\end{figure*}

\noindent\textbf{Different Dimensions of Base Functions.}
We test different dimensions of the \textit{base functions} $\mathbf{f}_k$ in Eqn.~\ref{eqa:base_function}. We set the dimension number $N$ from $16$ to $512$ while keeping other settings the same. Our batch size to $2$ and learning rate is $1e-4$. We only add $2$ raw position layers in these experiments for comparison. Each epoch contains $6k$ randomly chosen pictures in the NYUDepthv2 dataset. The evaluation results after each epoch are shown in Fig.~\ref{wlayer}. The bigger the number $N$ is, the faster and better the model converges. When $N=512$, the model performs the best. But for two reasons we can't increase the $N$ infinitely. Firstly the dimension number of the \textit{base functions} can not exceed the number of known sparse depth points, otherwise the Eqn.~\ref{eqa:lse} can't be solved. Secondly, as the dimension number of $\mathbf{f}_k$ increases, the least square function becomes harder to be solved, and it is more likely to meet numerical error. As a trade-off, we choose $N=128$ in the rest of our experiments.

\noindent\textbf{Positional Encoding.}
To prove that the position encoding part contributes to the result, we conduct an ablation study on positional encoding. We change the parameter $E$ in Eqn.~\ref{equ:pos_endode}. We set the batch size to $2$, learning rate to $1e-4$ and $N=128$. The results are shown in Fig.~\ref{pe}. \textit{No Position} means we don't use any positional information. \textit{Only position} means we only use the two raw positional layers which are the coordinates of the pixels in the image. \textit{Position encoding $x$} means we use positional encoding and set $E=x$. As we can see in the figure, with positional encoding layers, the model converges faster and better. But there is no big difference between different positional encoding layers. To make our model more efficient, we choose $E=5$ in the rest of the experiments.
 
 
\begin{figure*}[htb]
  \centering
  \includegraphics[width=0.8\textwidth]{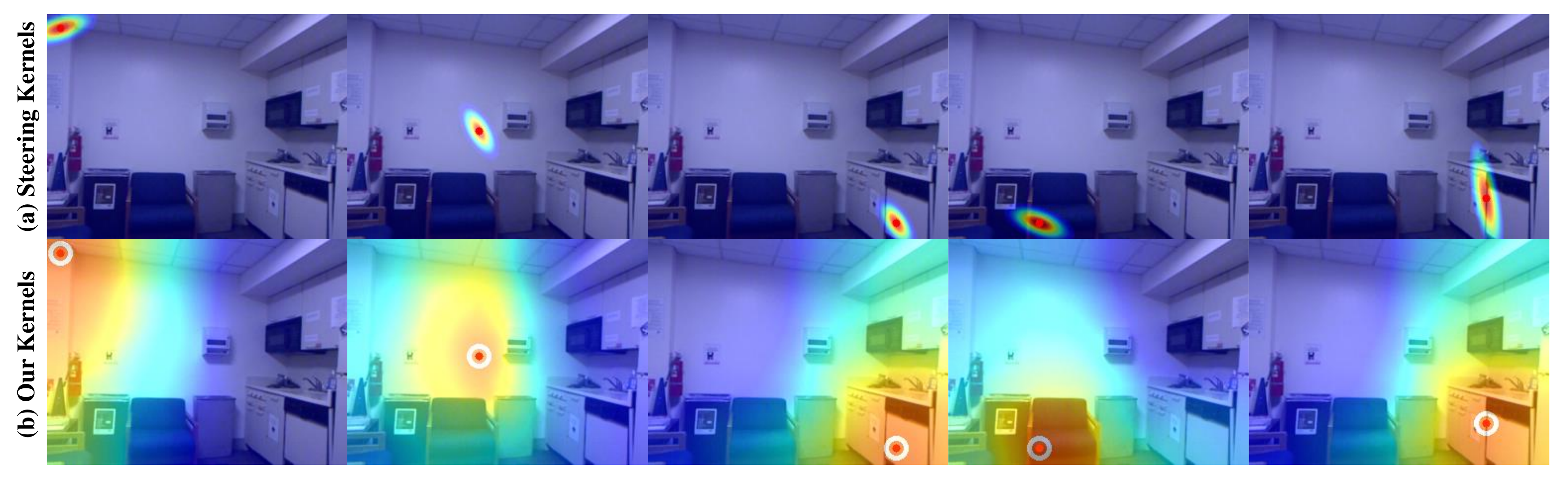}
  \caption{Kernel visualization comparison. The first line is the visualizations of the kernels of the Learning Steering Kernel model of some randomly sampled points in the picture. The second line is the visualizations of the kernels of our model of the same points in the picture.}
  \label{kernel_compare}
\end{figure*}

\subsection{Change of the Sparse Depth Maps}


\begin{figure*}
\label{REL_and_RMSE}
  \begin{minipage}{0.5\linewidth}
    \centering
    \includegraphics[scale=0.4]{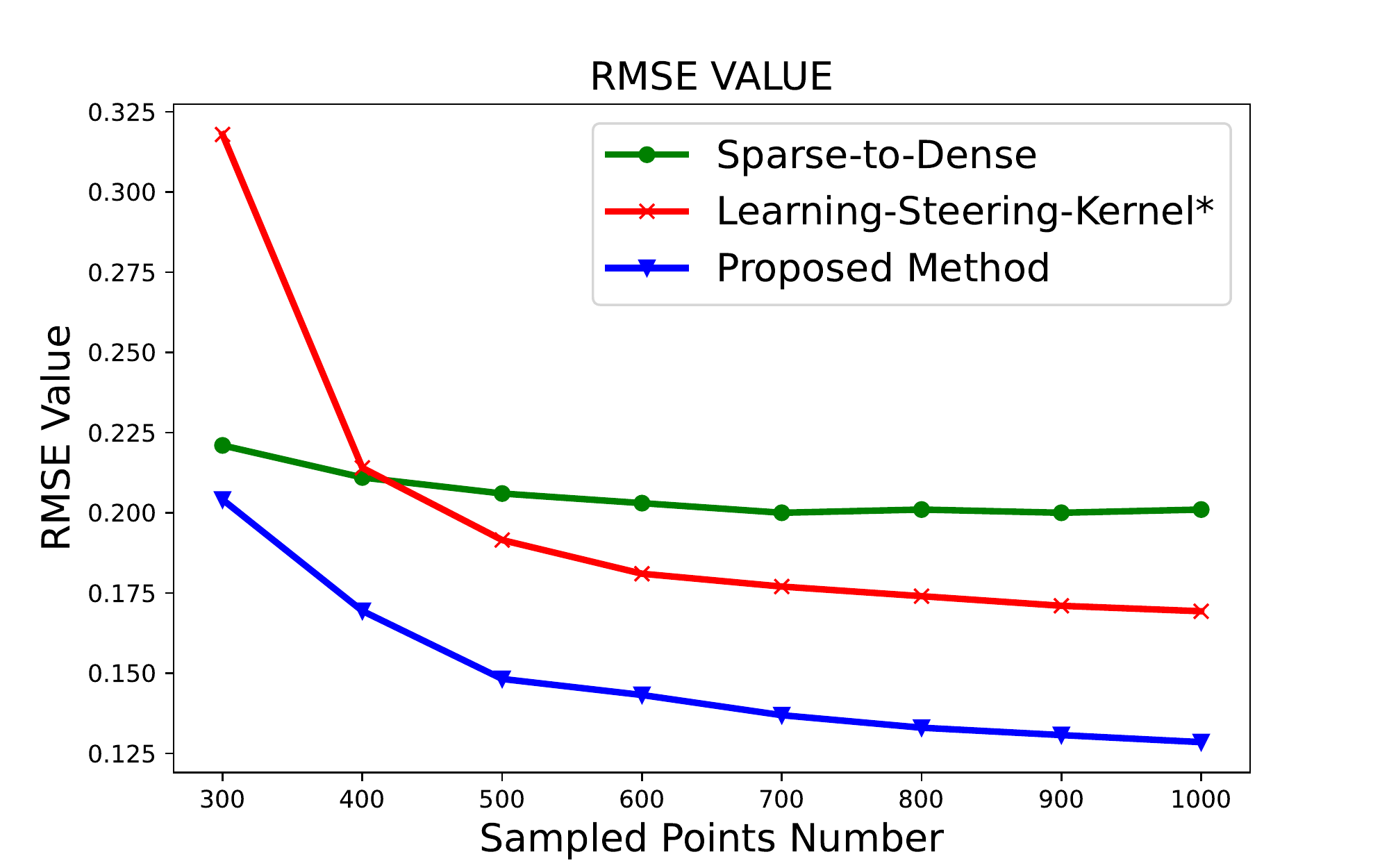}
  \end{minipage}%
  \begin{minipage}{0.5\linewidth}
    \centering
    \includegraphics[scale=0.35]{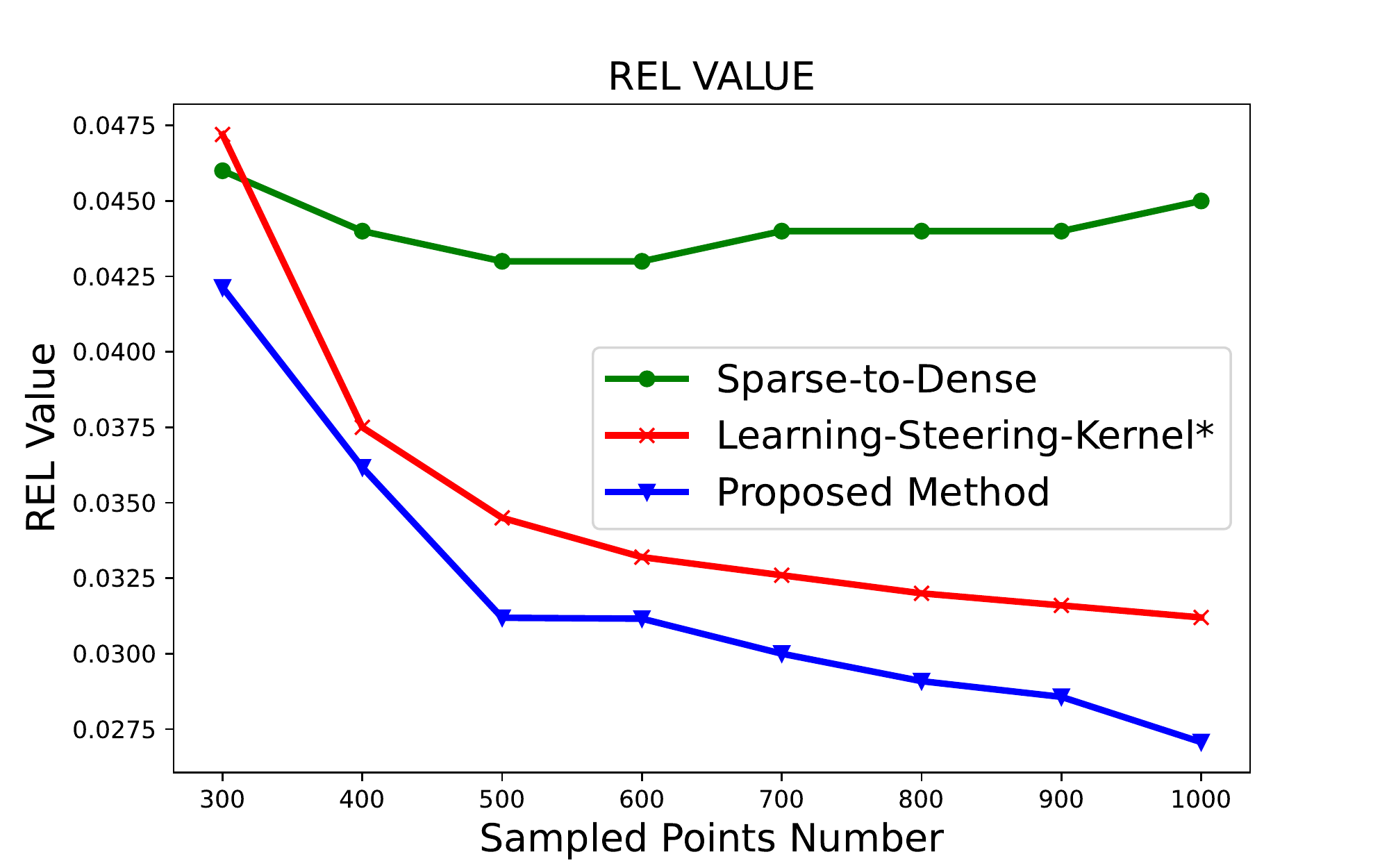}
  \end{minipage}
\caption{RMSE and REL Comparison between different methods on different sampled points number.}
\vspace{-10pt}
\end{figure*}

\begin{figure}[htb]
  \centering
  \includegraphics[width=0.4\textwidth]{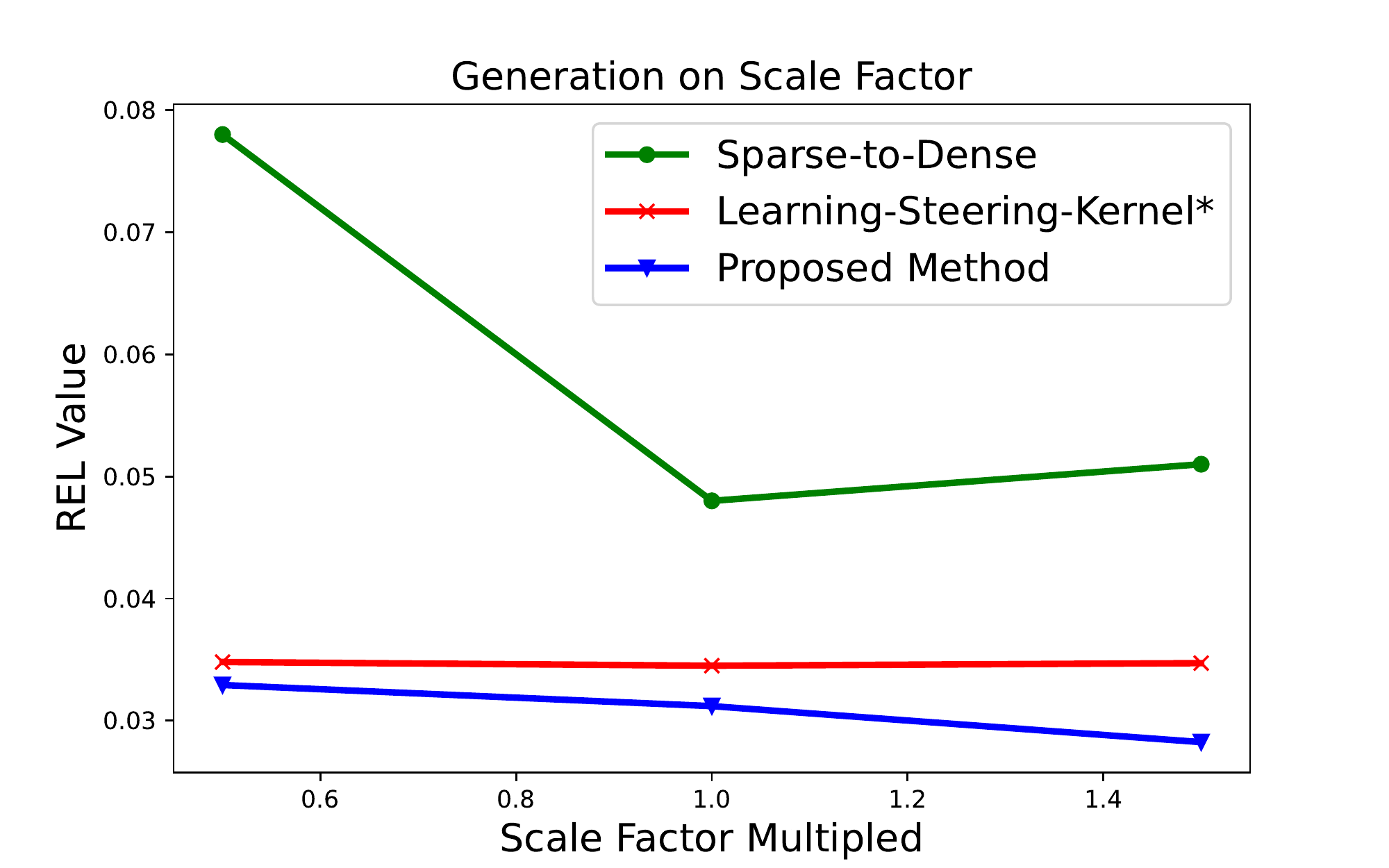}
  \caption{REL Comparison between different methods on different scale factor on the NYUDepthv2 Dataset.}
  \label{scale}
  \vspace{-10pt}
\end{figure}

\noindent\textbf{Different Numbers of Known Depth Measurements.}
In our method, the sparse depth measurements are used for least square estimation after all the neural network processes. So when we change the number of known depth measurements, the performance of the neural networks will not change at all. It will only influence the least square estimation process. With more known depth measurements, the final result ought to be better. We change the numbers of sampled points of known depth measurements and evaluate the models on the NYUDepthv2. The \textit{RMSE} and \textit{REL} results are shown in Fig.~\ref{REL_and_RMSE}. As the numbers of known depth measurements increase, the results of \textit{S2D} don't change much, while \textit{LSK}'s and our method's results increase obviously.
    
\noindent\textbf{Different Scale-Factors of the Sparse Depth Measurements.}
SLAM applications can extract the sparse points from images and measure their depth values, but the measurements may have a scale-factor error with the ground truth. In our method, the relationship between \textit{base functions} and the sparse depth values is a linear relationship. Therefore the dense depth map we predict ought to have the same scale-factor error as the input sparse depth measurements. This feature is useful in the future correction of the predicted depth values. We manually multiply the sampled sparse depth values by a factor and evaluate the models on NYUDepthv2. The \textit{REL} results are shown in Fig.~\ref{scale}. The \textit{REL} values of \textit{LSK} method and ours nearly stay still while \textit{S2D} suffers a lot.
    


\noindent\textbf{Sparse Points with Noise.}
In real-world applications, the sparse depth measurements may have noise. Because our method is based on the least square algorithm, it is built to reduce the impact of the noise. We can use \textit{IRLS} algorithm to decrease the influence of the noise of the sparse depth input. To prove that, we randomly choose $300$ points of the total $1000$ known depth points and add noise that is uniformly distributed in $[0,1]$. Only can our method apply \textit{IRLS} algorithm. Results are shown in Tab.~\ref{noise_table}. Noise obviously impacts the results of all these methods, and our method is affected the most. With \textit{IRLS} algorithm applied, we can reduce the impact of the noise.

\begin{table}[htb]
\centering
\begin{threeparttable}[b]
\caption{Evaluation Result of Noise Experiment}
\label{noise_table}

\begin{tabular}{c@{\ \ \ }c@{\ \ \ }c@{\ \ \ }c@{\ \ \ }c}
 \toprule
\multirow{2}{*}{Method} &  \multicolumn{2}{c}{Error($\downarrow$)} & \multicolumn{2}{c}{Change} \\
\cmidrule(lr){2-3}
\cmidrule(lr){4-5}
            & \multirow{2}{*}{RMSE}   &  \multirow{2}{*}{REL}     & RMSE      & REL       \\
            &        &          &   change  &  change   \\
 \midrule
 S2D        & 0.201  & 0.045    & -         &-          \\
 S2D+Noise  & 0.346  & 0.107    & $+0.145$  &$+6.20\%$  \\
 \midrule
 LSK         &0.169   &0.031     & -         &-          \\
 LSK+Noise   &0.358   &0.114     &$+0.189$   &$+8.30\%$  \\
 \midrule
 Ours       &  0.129 &  0.027& - & -   \\
 Ours+Noise &  0.366 & 0.120& $+0.237$&$+9.33\%$  \\
 Ours+Noise+IRLS& 0.331& 0.089 &$+0.203$&$+6.19\%$\\
 \bottomrule
\end{tabular}

\end{threeparttable}
\vspace{-0.5cm}
\end{table}
    
\subsection{Super-Resolution Testing}

\noindent\textbf{Kernel Visualization Comparison.}
In the Learning Steering Kernel algorithm \cite{liu2021learning}, which is also a depth-independent algorithm, each pixel has a regression kernel predicted from the RGB image. This kernel measures the affinity between other pixels and the chosen pixel. In our method, we have the base function of each pixel estimated from the RGB image. We calculate the correlation coefficient between the exact pixels and other pixels by using dot product to draw kernels as well. The result is shown in Fig.~\ref{kernel_compare}. We can see that the kernels in our method are more flexible and can cover the whole image while the kernels in the Learning Steering Kernel method can only be oval-like and only concern the local feature.

\begin{figure}
  \centering
  \includegraphics[width=0.45\textwidth]{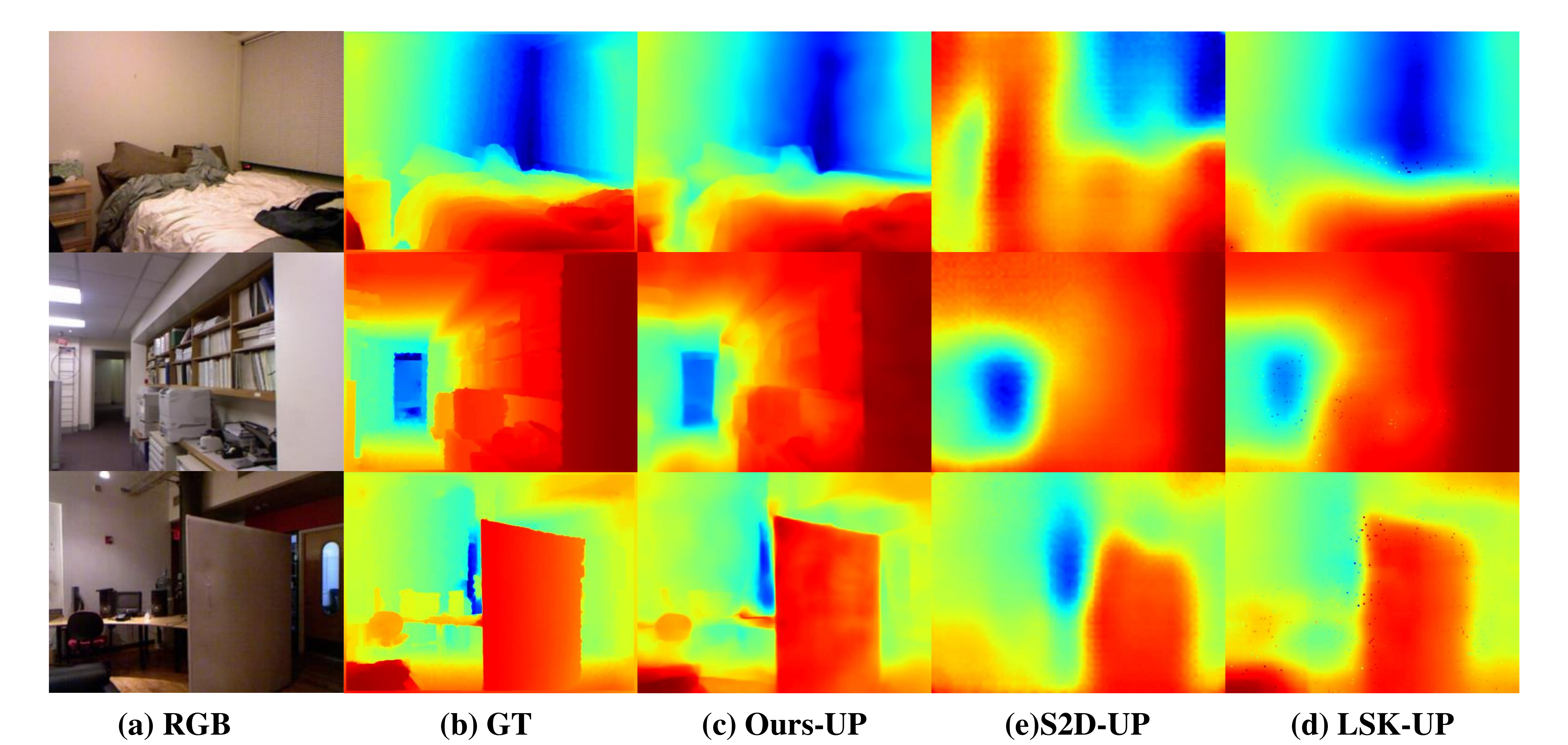}
  \caption{Super resolution results comparison. (a) is ground truth downsampled to $360\times480$. Others are upsampled to $360\times480$ by different methods.}
  \label{upsample_compare}
  \vspace{-10pt}
\end{figure}

\noindent\textbf{Super-Resolution Results Comparison.}
With the help of the two-stage strategy in our method, we can upsample the dense depth map in a reliable way. After feature extraction process, we get a feature map of the RGB image $ \Phi(\mathbf{c}) $. We first interpolate the feature map to our target resolution and then add the positional encoding layers which are rebuilt according to the new resolution into the feature map. Then we follow the normal step of our method. In the experiment, we first downsample the RGB image to $ 120\times160 $ and get its associated sparse depth map as the input. Then we upsample the dense depth map to $ 360\times480 $. For \textit{S2D} and \textit{LSK} algorithm, we simply upsample their output dense depth map using bi-linear interpolation. Besides \textit{RMSE} and \textit{REL} error, we also evaluate the \textit{Soft Edge Error (SEE)} metric on pixels belonging to object boundaries, defined as the minimum absolute error between the predicted depth map and the ground truth, according to \cite{Chen_2019_ICCV}. The visualization is shown in Fig.~\ref{upsample_compare} and the evaluation results are shown in Tab.~\ref{upsample_table}. The edges of upsampled dense depth maps predicted by our method are more clear, and the errors are smaller.

\begin{table}[htb]
\centering
\begin{threeparttable}[b]
\caption{Evaluation Result of Upsample Experiment}
\label{upsample_table}
\begin{tabular}{c@{\ \ \ \ \ \ \ }c@{\ \ \ \ \ \ \ }c@{\ \ \ \ \ \ \ }c}
 \toprule
\multirow{2}{*}{Method\tnote{1}} &  \multicolumn{3}{c}{Error($\downarrow$)} \\
\cmidrule(lr){2-4}
 & RMSE &  REL &  SEE  \\
 \midrule
 S2D&  0.487&  1.367&  0.901  \\
 LSK\tnote{2}&  0.059&  0.251&  0.056  \\
 Ours& \textbf{0.046}& \textbf{0.215}& \textbf{0.041} \\
 \bottomrule
\end{tabular}
  \begin{tablenotes}
     \item[1] \footnotesize{All take 1000 sampled points.}
     \item[2] \footnotesize{The result of Learning Steering Kernel part.}
  \end{tablenotes}
\end{threeparttable}
\vspace{-10pt}
\end{table}

\subsection{Generalization wrt. Different Datasets}

\begin{figure}
  \centering
  \includegraphics[width=0.45\textwidth]{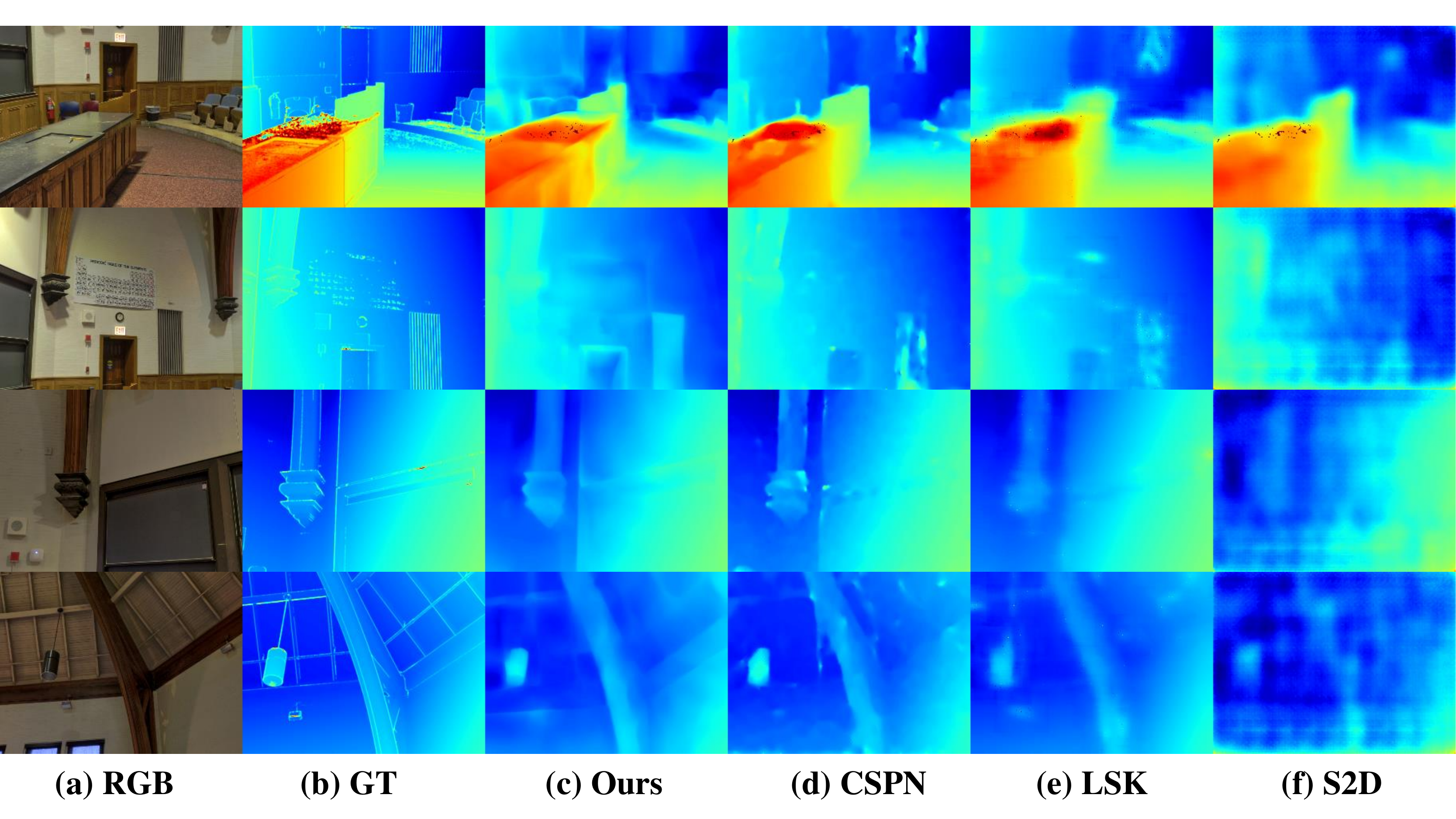}
  \caption{Generalization experiments on Diode dataset. In all methods the sampled number in sparse depth map is 1000.}
  \label{generation_diode}
  \vspace{-10pt}
\end{figure}

To test the generalization ability of our method, we validate the model trained on the NYUDepthv2 dataset directly on a new dataset DIODE (Dense Indoor and Outdoor DEpth). DIODE is a dataset that contains diverse high-resolution color images with accurate, dense, and far-range depth measurements. It is the first public dataset to include RGBD images of indoor and outdoor scenes obtained with one sensor suite. There are $325$ images in the indoor validation subset and $446$ images in the outdoor validation subset. In this part, we also test the SOTA algorithm CSPN method \cite{cheng2018depth}. The evaluation data is shown in Tab.~\ref{diode_indoor_table}, and the visualization result is shown in Fig.~\ref{generation_diode}. 

Our model contains a convolutional neural network and a fully connected neural network. The structures of both are simple and straight. We compare the parameter numbers of models of different methods. We also compare the runtimes of all these methods, using the same resolution $228\times304$. Comparison result is shown in Tab.~\ref{model_parameters}. Our method is the most light-weighted and fast.

\begin{table}[htb]
\centering
\begin{threeparttable}[b]
\caption{Generation Result on Diode Indoor Dataset}
\label{diode_indoor_table}

\begin{tabular}{c@{}c@{\ \ \ }c@{\ \ \ }c@{\ \ \ }c@{\ \ \ }c@{\ \ \ }c}
 \toprule
\multirow{2}{*}{Subset}&\multirow{2}{*}{Method\tnote{1}} &  \multicolumn{2}{c}{Error($\downarrow$)}  & \multicolumn{3}{c}{Accuracy($\uparrow$)} \\
\cmidrule(lr){3-4}
\cmidrule(lr){5-7}
  &&RMSE&  REL&  $ \delta_{1.25} $ & $ \delta_{1.25^2} $ & $ \delta_{1.25^3} $  \\
 \midrule
 \multirow{4}{*}{Indoor} & S2D&  0.462&  0.094&  95.50&  98.34&  99.19 \\
 &LSK\tnote{2}& 0.388&  0.061&  96.23&  98.71& 99.41 \\
  &CSPN & 0.292& \textbf{0.050}&  \textbf{97.52}&  \textbf{99.08}& \textbf{99.55} \\
 &Ours& \textbf{0.289}& 0.061& 96.55 & 98.89 & 99.48\\
 \midrule
  \multirow{4}{*}{Outdoor}&S2D  &  7.648&  1.827&  57.00&  70.40&  78.36 \\
   &LSK\tnote{2} &  5.361&  2.520&  71.68&  82.23& 87.33 \\
  &CSPN & 4.671& 2.665&  \textbf{76.64}&  \textbf{84.56}& \textbf{88.44} \\
 &Ours& \textbf{4.227}& \textbf{1.827}& 71.32 & 82.73 & 87.74\\
 \bottomrule
\end{tabular}

  \begin{tablenotes}
     \item[1] \footnotesize{All take 1000 sampled points.}
     \item[2] \footnotesize{The result of Learning Steering Kernel part.}
  \end{tablenotes}

\end{threeparttable}
\vspace{-5pt}
\end{table}

\begin{table}[htb]
\centering
\begin{threeparttable}[b]
\caption{Network Parameters}
\label{model_parameters}

\begin{tabular}{c@{\ \ \ \ \ \ }c@{\ \ \ \ \ \ }c}
 \toprule
 Method & Network Parameters($\downarrow$) & Runtime($\downarrow$)\\
 \midrule
 S2D&  63.566 M & 22ms\\
 LSK\tnote{1} & 13.422 M&114ms \\
  CSPN & 256.078 M & 67ms\\
 Ours& \textbf{3.466 M}&\textbf{15ms}\\
 \bottomrule
\end{tabular}
  \begin{tablenotes}
     \item[1] \footnotesize{Parameters in LSK part.}
  \end{tablenotes}
\end{threeparttable}
\vspace{-10pt}
\end{table}

\section{Conclusions}
\label{sec:conclusion}
In this paper, we propose a depth-independent least square estimation network to complete the sparse depth map to a dense one. We decouple the RGB image and the sparse depth map because the neural network in our model only extract the features of the RGB image. With experiments on the public datasets, we show that the performance of our method is promising both in accuracy and runtime. Besides, our model has a good performance when some features of the sparse depth map change. Moreover, our model can generalize well compared with other methods.

\printbibliography

@article{hu2022deep,
  title={Deep Depth Completion: A Survey},
  author={Hu, Junjie and Bao, Chenyu and Ozay, Mete and Fan, Chenyou and Gao, Qing and Liu, Honghai and Lam, Tin Lun},
  journal={arXiv preprint arXiv:2205.05335},
  year={2022}
}

@article{liu2021learning,
  title={Learning steering kernels for guided depth completion},
  author={Liu, Lina and Liao, Yiyi and Wang, Yue and Geiger, Andreas and Liu, Yong},
  journal={IEEE Transactions on Image Processing},
  volume={30},
  pages={2850--2861},
  year={2021},
  publisher={IEEE}
}

@article{takeda2007kernel,
  title={Kernel regression for image processing and reconstruction},
  author={Takeda, Hiroyuki and Farsiu, Sina and Milanfar, Peyman},
  journal={IEEE Transactions on image processing},
  volume={16},
  number={2},
  pages={349--366},
  year={2007},
  publisher={IEEE}
}

@inproceedings{ma2018sparse,
  title={Sparse-to-dense: Depth prediction from sparse depth samples and a single image},
  author={Ma, Fangchang and Karaman, Sertac},
  booktitle={2018 IEEE international conference on robotics and automation (ICRA)},
  pages={4796--4803},
  year={2018},
  organization={IEEE}
}

@inproceedings{bhutani2020unsupervised,
  title={Unsupervised Depth and Confidence Prediction from Monocular Images using Bayesian Inference},
  author={Bhutani, Vishal and Vankadari, Madhu and Jha, Omprakash and Majumder, Anima and Kumar, Swagat and Dutta, Samrat},
  booktitle={2020 IEEE/RSJ International Conference on Intelligent Robots and Systems (IROS)},
  pages={10108--10115},
  year={2020},
  organization={IEEE}
}

@inproceedings{cheng2018depth,
  title={Depth estimation via affinity learned with convolutional spatial propagation network},
  author={Cheng, Xinjing and Wang, Peng and Yang, Ruigang},
  booktitle={Proceedings of the European Conference on Computer Vision (ECCV)},
  pages={103--119},
  year={2018}
}

@inproceedings{park2020non,
  title={Non-local spatial propagation network for depth completion},
  author={Park, Jinsun and Joo, Kyungdon and Hu, Zhe and Liu, Chi-Kuei and So Kweon, In},
  booktitle={Computer Vision--ECCV 2020: 16th European Conference, Glasgow, UK, August 23--28, 2020, Proceedings, Part XIII 16},
  pages={120--136},
  year={2020},
  organization={Springer}
}

@inproceedings{saxena2005learning,
  title={Learning depth from single monocular images},
  author={Saxena, Ashutosh and Chung, Sung H and Ng, Andrew Y and others},
  booktitle={NIPS},
  volume={18},
  pages={1--8},
  year={2005}
}

@article{saxena2008make3d,
  title={Make3d: Learning 3d scene structure from a single still image},
  author={Saxena, Ashutosh and Sun, Min and Ng, Andrew Y},
  journal={IEEE transactions on pattern analysis and machine intelligence},
  volume={31},
  number={5},
  pages={824--840},
  year={2008},
  publisher={IEEE}
}

@inproceedings{vandenhende2020mti,
  title={Mti-net: Multi-scale task interaction networks for multi-task learning},
  author={Vandenhende, Simon and Georgoulis, Stamatios and Van Gool, Luc},
  booktitle={European Conference on Computer Vision},
  pages={527--543},
  year={2020},
  organization={Springer}
}

@inproceedings{xu2018pad,
  title={Pad-net: Multi-tasks guided prediction-and-distillation network for simultaneous depth estimation and scene parsing},
  author={Xu, Dan and Ouyang, Wanli and Wang, Xiaogang and Sebe, Nicu},
  booktitle={Proceedings of the IEEE Conference on Computer Vision and Pattern Recognition},
  pages={675--684},
  year={2018}
}

@article{czarnowski2020deepfactors,
  title={Deepfactors: Real-time probabilistic dense monocular slam},
  author={Czarnowski, Jan and Laidlow, Tristan and Clark, Ronald and Davison, Andrew J},
  journal={IEEE Robotics and Automation Letters},
  volume={5},
  number={2},
  pages={721--728},
  year={2020},
  publisher={IEEE}
}

@article{qin2018vins,
  title={Vins-mono: A robust and versatile monocular visual-inertial state estimator},
  author={Qin, Tong and Li, Peiliang and Shen, Shaojie},
  journal={IEEE Transactions on Robotics},
  volume={34},
  number={4},
  pages={1004--1020},
  year={2018},
  publisher={IEEE}
}

@article{mur2015orb,
  title={ORB-SLAM: a versatile and accurate monocular SLAM system},
  author={Mur-Artal, Raul and Montiel, Jose Maria Martinez and Tardos, Juan D},
  journal={IEEE transactions on robotics},
  volume={31},
  number={5},
  pages={1147--1163},
  year={2015},
  publisher={IEEE}
}

@inproceedings{chang2018pyramid,
  title={Pyramid stereo matching network},
  author={Chang, Jia-Ren and Chen, Yong-Sheng},
  booktitle={Proceedings of the IEEE Conference on Computer Vision and Pattern Recognition},
  pages={5410--5418},
  year={2018}
}

@article{liu2018intriguing,
  title={An intriguing failing of convolutional neural networks and the coordconv solution},
  author={Liu, Rosanne and Lehman, Joel and Molino, Piero and Such, Felipe Petroski and Frank, Eric and Sergeev, Alex and Yosinski, Jason},
  journal={arXiv preprint arXiv:1807.03247},
  year={2018}
}

@article{liu2017learning,
  title={Learning affinity via spatial propagation networks},
  author={Liu, Sifei and De Mello, Shalini and Gu, Jinwei and Zhong, Guangyu and Yang, Ming-Hsuan and Kautz, Jan},
  journal={Advances in Neural Information Processing Systems},
  volume={30},
  year={2017}
}

@InProceedings{Chen_2019_ICCV,
author = {Chen, Chuangrong and Chen, Xiaozhi and Cheng, Hui},
title = {On the Over-Smoothing Problem of CNN Based Disparity Estimation},
booktitle = {Proceedings of the IEEE/CVF International Conference on Computer Vision (ICCV)},
month = {October},
year = {2019}
}

@inproceedings{Silberman:ECCV12,
  author    = {Nathan Silberman, Derek Hoiem, Pushmeet Kohli and Rob Fergus},
  title     = {Indoor Segmentation and Support Inference from RGBD Images},
  booktitle = {ECCV},
  year      = {2012}
}

@article{ming2021deep,
  title={Deep learning for monocular depth estimation: A review},
  author={Ming, Yue and Meng, Xuyang and Fan, Chunxiao and Yu, Hui},
  journal={Neurocomputing},
  volume={438},
  pages={14--33},
  year={2021},
  publisher={Elsevier}
}

@article{huang2019hms,
  title={Hms-net: Hierarchical multi-scale sparsity-invariant network for sparse depth completion},
  author={Huang, Zixuan and Fan, Junming and Cheng, Shenggan and Yi, Shuai and Wang, Xiaogang and Li, Hongsheng},
  journal={IEEE Transactions on Image Processing},
  volume={29},
  pages={3429--3441},
  year={2019},
  publisher={IEEE}
}

@inproceedings{zhang2018deep,
  title={Deep depth completion of a single rgb-d image},
  author={Zhang, Yinda and Funkhouser, Thomas},
  booktitle={Proceedings of the IEEE Conference on Computer Vision and Pattern Recognition},
  pages={175--185},
  year={2018}
}

@InProceedings{Qu_2020_WACV,
author = {Qu, Chao and Nguyen, Ty and Taylor, Camillo},
title = {Depth Completion via Deep Basis Fitting},
booktitle = {Proceedings of the IEEE/CVF Winter Conference on Applications of Computer Vision (WACV)},
month = {March},
year = {2020}
}

@inproceedings{jaritz2018sparse,
  title={Sparse and dense data with cnns: Depth completion and semantic segmentation},
  author={Jaritz, Maximilian and De Charette, Raoul and Wirbel, Emilie and Perrotton, Xavier and Nashashibi, Fawzi},
  booktitle={2018 International Conference on 3D Vision (3DV)},
  pages={52--60},
  year={2018},
  organization={IEEE}
}

@INPROCEEDINGS{8793637,
  author={Ma, Fangchang and Cavalheiro, Guilherme Venturelli and Karaman, Sertac},
  booktitle={2019 International Conference on Robotics and Automation (ICRA)}, 
  title={Self-Supervised Sparse-to-Dense: Self-Supervised Depth Completion from LiDAR and Monocular Camera}, 
  year={2019},
  volume={},
  number={},
  pages={3288-3295},
  doi={10.1109/ICRA.2019.8793637}}

@ARTICLE{8869936,
  author={Cheng, Xinjing and Wang, Peng and Yang, Ruigang},
  journal={IEEE Transactions on Pattern Analysis and Machine Intelligence}, 
  title={Learning Depth with Convolutional Spatial Propagation Network}, 
  year={2020},
  volume={42},
  number={10},
  pages={2361-2379},
  doi={10.1109/TPAMI.2019.2947374}}

@inproceedings{cheng2020cspn++,
  title={Cspn++: Learning context and resource aware convolutional spatial propagation networks for depth completion},
  author={Cheng, Xinjing and Wang, Peng and Guan, Chenye and Yang, Ruigang},
  booktitle={Proceedings of the AAAI Conference on Artificial Intelligence},
  volume={34},
  number={07},
  pages={10615--10622},
  year={2020}
}

@article{tang2018ba,
  title={Ba-net: Dense bundle adjustment network},
  author={Tang, Chengzhou and Tan, Ping},
  journal={arXiv preprint arXiv:1806.04807},
  year={2018}
}

\end{document}